\documentclass[11pt,a4paper]{article}
\usepackage[hyperref]{acl2017}
\usepackage{times}
\usepackage{latexsym}
\usepackage{graphicx}

\usepackage{url}

\aclfinalcopy 

\title{A Survey of Question Answering for Math and Science Problem}

\author{Arindam Bhattacharya \\
  Indian Institute of Technology, Delhi \\
  {\tt arindam@cse.iitd.ac.in}
}

\date{}

\begin{document}
\maketitle
\begin{abstract}
  Turing test was long considered the measure for artificial intelligence. But with the advances in AI, it has proved to be insufficient measure. We can now aim to measure machine intelligence like we measure human intelligence. One of the widely accepted measure of intelligence is standardized math and science test.

  In this paper, we explore the progress we have made towards the goal of making a machine smart enough to pass the standardized test. We see the challenges and opportunities posed by the domain, and note that we are quite some ways from actually making a system as smart as a even a middle school scholar.
\end{abstract}

\section{Introduction}

Humans intelligence is often measured by asking them questions and evaluating their answers. Starting from infancy, through school, college and career, humans are judged by evaluating answers that they give to questions posed to them. Such an approach is so widespread that it is possible for someone to believe that it is the only way to measure intelligence. Yet, for long we have tried to quantify artificial intelligence in a very different way.

The Turing test, proposed by Alan Turing in 1950 states that: if a system can exhibit conversational behaviour that is indistinguishable from that of a human during a conversation, that system could be considered intelligence \citep{Turing:50}. Since then it has been often critisized of being a poor measure for the purpose \citep{Hayes:95}. It has also since been proposed that standardized tests in mathemetics and science could be a suitable measure for judging machine intelligence \citep{Clark:16}. To this end, there has been attempts to develop Question Answering systems aimed at the task of solving standardized tests of math and science problems.

In this paper, we will briefly explore the various dimesions along which attempts have been made to tackle both math and science problems. For mathematics, we will explore systems that tries to solve algebraic word problems and geometry problems. For science, we will explore systems that solves questions ranging from basic questions (e.g. definitional knowledge) to questions that require complex world knowledge, and even diagrams.

\section{Question Answering and the Math/Science domain}

Question Answering (QA) is the task of generating or extracting an answer to a question posed in natural language. Modern QA systems can be divided into two broad paradigms \citep{Jurafsky:16}. The first paradigm, called text-based (or IR based) QA relies on large volumes of text. Given a question, it uses information retrieval methods on the available corpus and retrieves documents which contains the answer. The candidate answers are then extracted from the text and ranked. The second paradigm is knowledge based QA, where we create a semantic representation of the question and use it to query databases of facts.

Focusing on science and mathematics as domain for QA presents certain unique challanges. Solving these problems require not only understanding question for which we levarage language processing, but also to maintain an internal representation of the problem and often carry out symbolic computation \citep{Clark:16}. They cannot be easily answered by information retrieval or knowledge based methods.

Along with the challanges, the domain also provides us with certain unique oppotunities. Using standardized test provides us with questions that are graduated by difficulty and multifaceted in nature: different questions explore different types of knowledge \citep{Carissa:16}.

\section{Question Answering for Science}
This section will focus on Science questions of standardized tests. The types of quesions include basic fact retrieval, inference and world knowlege, and diagrams. We also explore the state of the art in these types of quesions.

\subsection{Dataset}
\label{sec:dataset}

The standardized test for Science used for the QA tasks is the New York Regents Science Exams \citep{Regent:14}. Following are examples of types of questions in the test:

\subsubsection*{Basic Questions}

These questions are factoid type questions requiring definitional knowledge. Example:
\begin{enumerate}
\item Which object is the best conductor of electricity? (A) a wax crayon (B) a plastic spoon (C) a rubber eraser (D) an iron nail
\item The movement of soil by wind or water is called (A) condensation (B) evaporation (C) erosion (D) friction
\end{enumerate}
An IR based QA system can address these questions.

\subsubsection*{Simple Inference}
\label{sec:simple-inference}

These are questions which require simple inference over known facts to arrive at the answer. Example:
\begin{enumerate}
\item Which example describes an organism taking in nutrients? (A) dog burying a bone (B) A girl eating an apple (C) An insect crawling on a leaf (D) A boy planting tomatoes in the garden
\end{enumerate}
Answering this question requires knowledge that eating involves taking in nutrients, and that an apple contains nutrients.

\subsubsection*{More Complex World Knowledge}
\label{sec:more-complex-world}

These questions require deeper world knowledge and more advanced liguistic capabilities for the system to be able to understand questions and produce an answer. Example:

\begin{enumerate}
\item A student riding a bicycle observes that it moves faster on a smooth road than on a rough road. This happens because the smooth road has (A) less gravity (B) more gravity (C) less friction (D) more friction
\end{enumerate}
To answer this question the system must be aware of the fact that riding a bicycle implies moving it, and infer logically the path [smooth -> less friction -> faster movement].

\subsubsection*{Diagram}
\label{sec:diagram}

\begin{figure}[h]
  \label{fig:plant}
  \centering
  \includegraphics{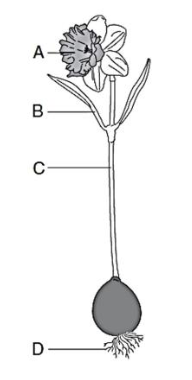}
  \caption{Parts of plant.}
\end{figure}

Questions with diagrams are quite common in these test. The diagrams includes sketchs, maps, graphs, tables etc. These often proves to be quite difficult for the QA systems. Example:
\begin{enumerate}
\item Which letter in the diagram \ref{fig:plant} points to the plant structure that takes in water and nutrients?
\end{enumerate}

\subsection{Models}
\label{sec:models}

Various approaches, ranging from a combination of Information Retrieval, Statistics and Inference to Integer Programming  are employed to tackle the aforementioned challanges.

\begin{figure}[ht]
  \centering
  \includegraphics[scale=.6]{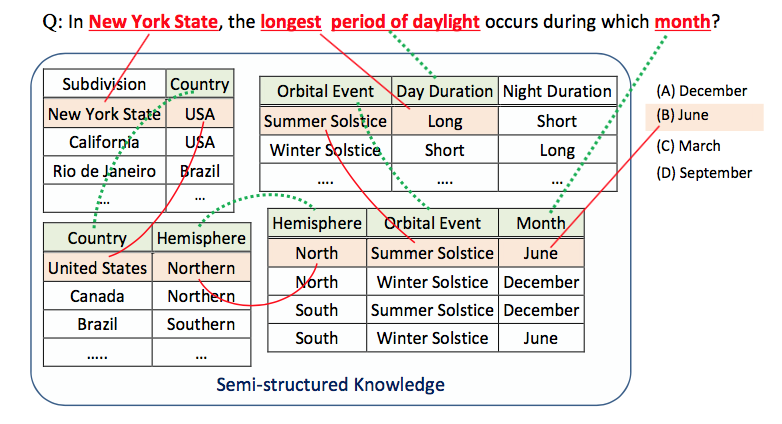}
  \caption{TableILP searches for the best support graph (chains of reasoning) connecting the question to an answer, in this case June.}
  \label{fig:ilp}
\end{figure}

\citep{Daniel:16} proposes a method for solving QA using Integer Linear Programming. Given semi-structured knowledge as tables, the QA problem is formulated as that of finding a desirable Support Graph (Fig. \ref{fig:ilp}), which in turn is formulated as ILP.

\begin{figure}[ht]
  \centering
  \includegraphics[scale=.6]{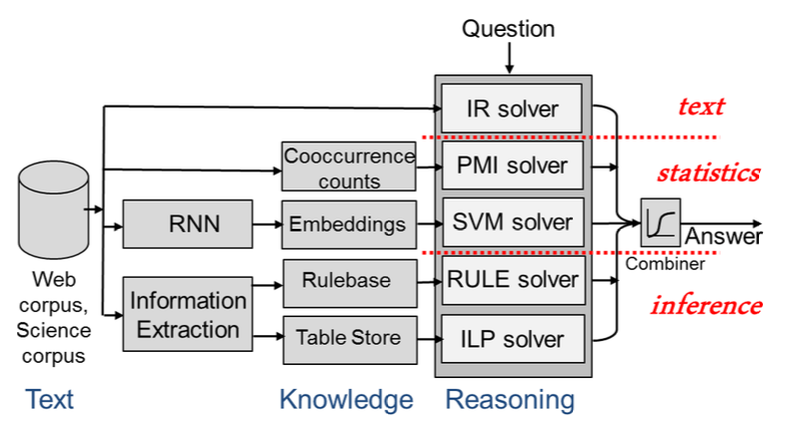}
  \caption{Aristo uses five solvers, each using different types of knowledge, to answer multiple choice questions.}
  \label{fig:aristo}
\end{figure}

The state of the art system, ARISTO \citep{Peter:16} employs an ensemble of solvers that tackle the problem at various layers. The layers are: Information Retrieval solver, Pointwise Mutual Information solver, Support Vector Machine solver, RULE solver that contains hand coded rules, and an Integer Linear Programming solver. The system is shown in Fig. \ref{fig:aristo}
On non-diagram, multiple choice science questions (NDMC), Aristo system currently scores on average 75\% (4th grade), 63\% (8th grade), and 41\% (12th grade) on (previously unseen) NY Regents Science Exams (NDMC questions only, typically 4-way multiple choice). As can be seen, questions become considerably more challenging at higher grade levels. On a broader multi-state collection of 4th grade NDMC questions, Aristo scores 65\% (unseen questions). Note that these are the "easier" questions (no diagrams, multiple choice); other question types pose additional challenges as we have described. There are no good systems that can tackle the questions out of NDMC and no system to date comes even close to passing a full 4th grade science exam.

\section{Question Answering for Mathematics}
\label{sec:quest-answ-math}

Math questions cannot be solved by IR systems. The basic strategy in mathematics, especially arithmatic quesions, is to understand the problem and formulate an equation that can be calculated. Geometry questions poses difficulties because of their reliance on diagrams.

\subsection{Dataset}
For algebra questions, the standardized tests of \citep{Regent:14} is used. For geometry, questions from SATs are used.

\subsubsection*{Algebraic Problems}
\label{sec:algebraic-problems}

Algebraic problems are posed as stories that require language processions. Example:
\begin{enumerate}
\item Molly owns the Wafting Pie Company. This morning, her employees used 816 eggs to bake pumpkin pies. If her employees used a total of 1339 eggs today, how many eggs did they use in the afternoon?
\end{enumerate}
Some of them requires world modelling as well, for example:
\begin{enumerate}
\item Sara’s high school won 5 basketball games this year. They lost 3 games. How many games did they play in all?
\item John has 8 orange balloons, but lost 2 of them. How many orange balloons does John have now?
\end{enumerate}
Here, for the first, the knowledge that game is either won or lost is needed. The second example also has ``lost'' but results in a subtraction problem rather the addition like the first.

\subsubsection*{Geometry Problems}
\label{sec:geometry-problems}

\begin{figure}[h]
  \centering
  \includegraphics[scale=.7]{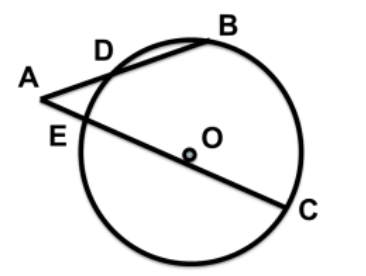}
  \caption{In the diagram, AB intersects circle O at D, AC intersects circle O at E, AE = 4, AC = 24, and AB = 16. Find AD.}
  \label{fig:geo}
\end{figure}

Geometry problems combine arithmatic and diagrammatic reasoning. Example: Fig.\ref{fig:geo}

\subsection{Models}
\label{sec:models-1}

\begin{figure}[h]
  \centering
  \includegraphics[scale=.6]{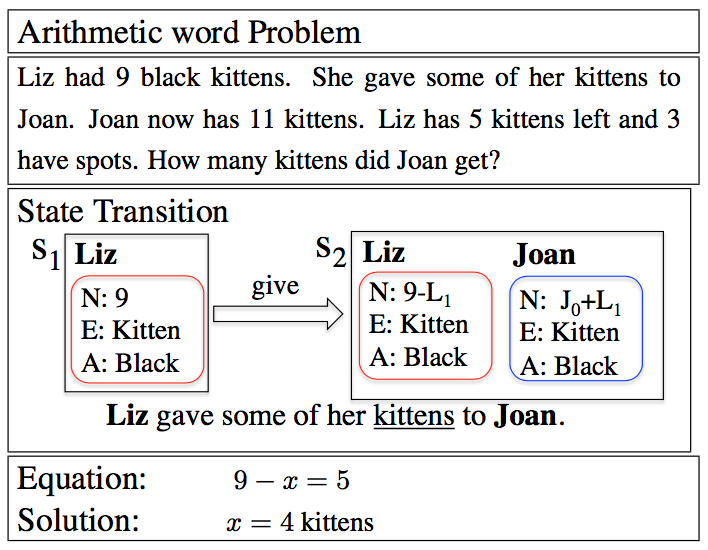}
  \caption{Verb Categorization}
  \label{fig:verb}
\end{figure}

One of the earliest attempt at solving algebraic word problems employed simple verb categorization \citep{Hoss:14}. The model extracted the verbs from the question and tried to formulate equations based on the verb category. The model is presented in Fig. \ref{fig:verb}.

\begin{figure}[h]
  \centering
  \includegraphics[scale=0.6]{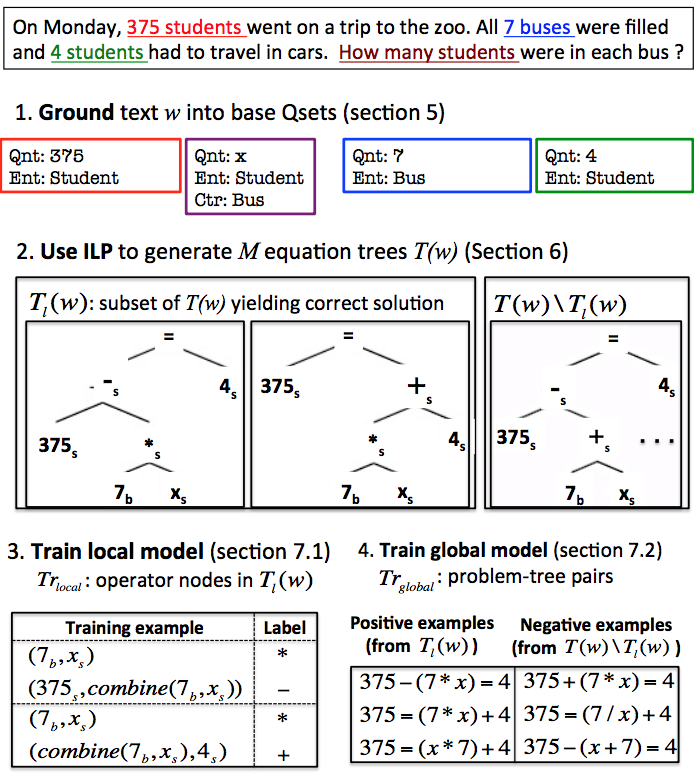}
  \caption{An overview of the process of learning for a word problem and its Qsets.}
  \label{fig:alges}
\end{figure}

A more sophisticated (and current state of the art) system, ALGES \citep{alges:15} uses Integer Linear Programming to map the word problems into equation trees. In contrast to \citep{Hoss:14}, which covered only $+,-$, ALGES covers $+,-,*,/$. They are also able to solve multi-sentence problem, unlike previous models. ALGES proceeds by extracting sets of ground truths from the text, and using ILP to form trees using them. Overview of the system is presented in Fig. \ref{fig:alges}.
ALGES produces an accuracy of 72\%, which is a 53\% error reduction over the previous best methods (verb categorization).

\begin{figure}[h]
  \centering
  \includegraphics[scale=.6]{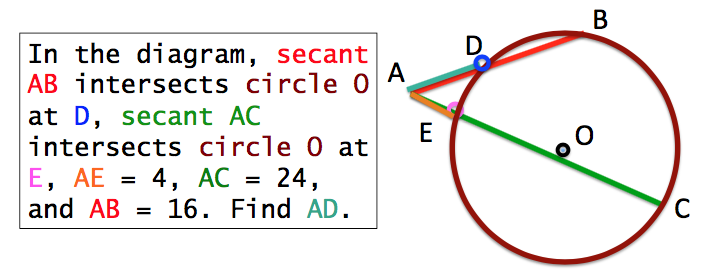}
  \caption{Diagram understanding: identifying visual elements in the diagram and aligning them with their textual mentions. Visual elements and their corresponding textual mentions are color coded. This Figure is best viewed in color.}
  \label{fig:galign}
\end{figure}

Geomertry quesions poses much greater difficulty, as expected, and the state of art systems can only achieve a score of 49\% in standard SAT. The initial work in this area aimed at aligning text with geometric diagrams \citep{galign:14}. It achieved the goal in three steps: identifying the primitives in the figure by picking elements that maximized the pixel coverage, agreement between the primitives and textual elements, and maximizing the coherence of the elements. Fig. \ref{fig:galign} shows the alignment achieved by the system.

\begin{figure}[h]
  \centering
  \includegraphics[scale=.6]{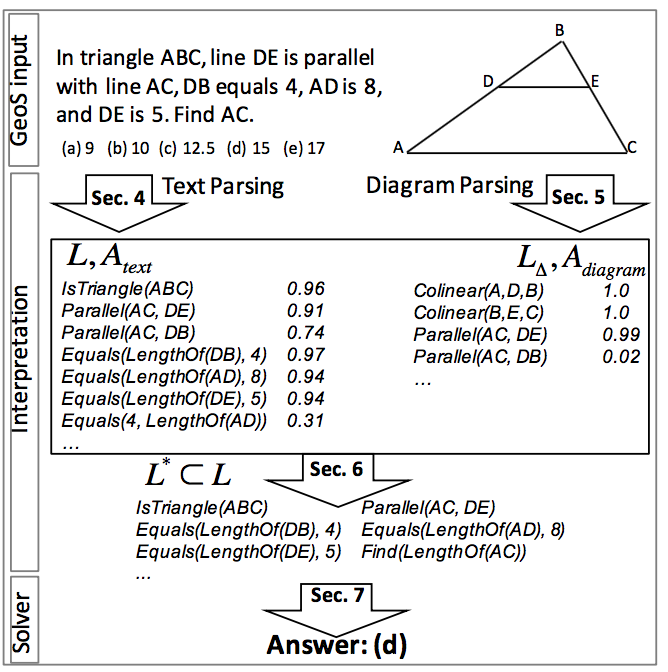}
  \caption{Overview of our method for solving geometry questions.}
  \label{fig:geos}
\end{figure}

While \citep{galign:14} understands the diagram, it could not solve geometric problems. GeoS \citep{geos:16} builds on the previous model and can answer geometric questions. It works in two steps: first it uses GAligner and language processing to convert the diagram and question into logical expressions, and then, it uses satisfiability solver to deduce the answer. Fig. \ref{fig:geos} summarizes the model.

\section{Conclusion}
\label{sec:conclusion}

We have seen that the current state of the art is not good at solving standardized tests (the smartest AI could not pass high school). There is a long way to go for AI. The field of QA systems on standardized math and science questions is fairly nascent with room for improvment with e.g. neural models, especially for diagram based questions. But the lack of results may also indicate that the current trend of data driven AI may not be the one stop solution to all problems.

We started the discussion with how Turing test is no longer the best measure for intelligence and agreed with \citep{Clark:16} that standardized tests would be a better judge of intelligence for machines, that is more in line with how we measure human intelligence. But if the system could pass the standardized test, would it be intelligent? \citep{weston2015towards} argues that standardized math and science tests can not test common sense, and hence not a true test for intelligence. Nonetheless, passing the test would be a landmark in the history of AI, as was the moment of passing Turing test.

\bibliography{acl2017}
\bibliographystyle{acl_natbib}

\end{document}